\documentclass[10pt, a4paper]{article}
\usepackage{lrec}
\usepackage{multibib}
\newcites{languageresource}{Language Resources}
\usepackage{graphicx,esvect}
\usepackage{tabularx}
\usepackage{soul}
\usepackage[linesnumbered,ruled]{algorithm2e} 
\usepackage{epstopdf}
\usepackage[utf8]{inputenc}

\usepackage{hyperref}
\usepackage{xstring}

\usepackage{color}

\title{Investigating an approach for low resource language dataset creation, curation and classification: Setswana and Sepedi}

\name{
    \parbox{\linewidth}{\centering
      Vukosi Marivate${}^{1}{}^{,}{}^{2}$, 
      Tshephisho Sefara$^{2}$, 
      Vongani Chabalala$^{3}$, 
      Keamogetswe Makhaya$^{4}$,
      Tumisho Mokgonyane$^{5}$, 
      Rethabile Mokoena$^{6}$, 
      Abiodun Modupe${}^{7}{}^{,}{}^{1}$%
    }%
  }

\address{University of Pretoria$^{1}$, CSIR$^{2}$, University of Zululand$^{3}$, University of Cape Town$^{4}$,\\ University of Limpopo$^{5}$, North-West University$^{6}$, University of the Witwatersrand$^{7}$ \\
         vukosi.marivate@cs.up.ac.za, tsefara@csir.co.za\\
         }

\abstract{
The recent advances in Natural Language Processing have been a boon for well represented languages in terms of available curated data and research resources. One of the challenges for low-resourced languages is clear guidelines on the collection, curation and preparation of datasets for different use-cases. In this work, we take on the task of creation of two datasets that are focused on news headlines (i.e short text) for Setswana and Sepedi and creation of a news topic classification task. We document our work and also present baselines for classification. We investigate an approach on data augmentation, better suited to low resource languages, to improve the performance of the classifiers.
}

\begin{document}

\maketitleabstract

\section{Introduction}

The most pressing issue with low-resource languages is of insufficient language resources. In this study, we introduce an investigation of a low-resource language that provides automatic formulation and customization of new capabilities from existing ones. While there are more than six thousand languages spoken globally, the openness of resources for each is extraordinarily unbalanced~\cite{nettle1998explaining}. For example, if we focus on language resources annotated on the public domain, as of November 2019, AG corpus released about $496,835$ news articles only in English languages from more than $200$ sources\footnote{\url{http://groups.di.unipi.it/~gulli}}, Reuters News Dataset~\cite{lewis1997reuters} comprise an roughly $10,788$ annotated texts from the Reuters financial newswire. The New York Times Annotated Corpus~\cite{sandhaus2008new} holds over $1.8$ million articles, and there are no standard annotated tokens in the low-resource language. Google Translate only supports around 100 languages \cite{johnson2017google}. A significant number of bits of knowledge focus on a small number of languages
neglecting $17\%$ out of the world’s language categories label as low-resource \cite{strassel2016lorelei}, which makes it challenging to develop various mechanisms for Natural Language Processing (NLP).


In South Africa 
several of the news websites (private and public) are published in English, even though there are 11 official languages (including English). We list the top premium newspapers by circulation as per first Quarter 2019~\cite{abc2019q1} in Table~\ref{tab:circulation}. We do not have a distinct collection of a diversity of languages with most of the reported datasets as existed in English, Afrikaans and isiZulu. In this work, we aim to provide a general framework that enables us to create an annotated linguistic resource for Setswana and Sepedi news headlines. We apply data sources of the news headlines from the South African Broadcast Corporation (SABC)~\footnote{\url{http://www.sabc.co.za/}}, their social media streams and a few acoustic news. Unfortunately, we do not have any direct access to news reports, so we hope this study will promote collaboration between the national broadcaster and NLP researchers.

\begin{table}[ht]
\begin{center}

\caption{Top newspapers in South Africa with their languages}
\label{tab:circulation}
\begin{tabular}{|l|l|l|}
\hline
Paper & Language & Circulation \\ \hline
Sunday Times & English & 260132  \\ \hline
Soccer Laduma & English &  252041 \\ \hline
Daily Sun & English & 141187 \\ \hline
Rapport & Afrikaans & 113636 \\ \hline
Isolezwe & isiZulu  & 86342 \\ \hline
Sowetan & English  & 70120 \\ \hline
Isolezwe ngeSonto & isiZulu & 65489 \\ \hline
Isolezwe ngoMgqibelo & isiZulu & 64676 \\ \hline
Son & Afrikaans & 62842 \\ \hline
\end{tabular}
\end{center}
\end{table}

The rest of the work is organized as follows. Section~\ref{sec:prior} discusses prior work that has gone into building local corpora in South Africa and how they have been used. Section~\ref{sec:approach} presents the proposed approach to build a local news corpora and annotating the corpora with categories. From here, we focus on ways to gather data for vectorization and building word embeddings (needing an expanded corpus). We also release and make pre-trained word embeddings for 2 local languages
as part of this work~\cite{vukosi_marivate_2020_3668481}. Section~\ref{sec:results} investigate building classification models for the Setswana and Sepedi news and improve those classifiers using a 2 step augmentation approach inspired by work on hierarchical language models \cite{yu2019hierarchical}. Finally, Section~\ref{sec:future} concludes and proposes a path forward for this work. 


\section{Prior Work}\label{sec:prior}

Creating sizeable language resources for low resource languages is important in improving available data for study \cite{zoph2016transfer} and cultural preservation. If we focus our attention on the African continent, we note that there are few annotated datasets that are openly available for tasks such as classification. In South Africa, the South African Center for Digital Language Resources (SADILAR)~\footnote{\url{www.sadilar.org}} has worked to curate datasets of local South African languages. There remain gaps such as accessing large corpora and data from sources such as broadcasters and news organizations as they have sizeable catalogs that are yet to make it into the public domain. In this work, we work to fill such a gap by collecting, annotating and training classifier models for news headlines in Setswana and Sepedi. As the data that we do find publicly is still small, we also have to deal with the challenges of Machine Learning on small data.

Machine learning systems perform poorly in presence of small training sets due to overfitting.
To avoid this problem, data augmentation can be used. The technique is well known in the field of image processing \cite{cubuk2019autoaugment}. Data augmentation refers to the augmentation of the training set with artificial, generated, training examples. This technique is used less frequently in NLP but a number of few studies applied data augmentation. 

\newcite{silfverberg2017data} use data augmentation to counteract overfitting where  recurrent neural network (RNN) Encoder-Decoder is implemented specifically geared toward a low-resource setting. Authors apply data augmentation by finding words that share word stem for example \textbf{fizzle} and \textbf{fizzling} share \textbf{fizzl}. Then authors replace a stem with another string. 

\newcite{zhang2015character} apply data augmentation by using synonyms as substitute words for the original words. However, \newcite{kobayashi2018contextual} states that synonyms are very limited and the synonym-based augmentation cannot produce numerous different patterns from the original texts. Hence, \newcite{kobayashi2018contextual} proposes contextual data augmentation by replacing words that are predicted by a language model given the context surrounding the original words to be augmented.

As \newcite{wei2019eda} states that these techniques are valid, they are not often used in
practice because they have a high cost of implementation relative to performance gain. They propose an easy data augmentation as techniques for boosting performance
on text classification tasks. These techniques involve synonym replacement, random insertion, random swap, and random deletion of a word. Authors observed good performance when using fraction of the dataset (\%):{1, 5, 10, 20, 30, 40, 50, 60, 70, 80, 90, 100}, as the data size increases, the accuracy also increases for augmented and original data. Original data obtained highest accuracy of 88.3\% at 100\% data size while augmented data obtained accuracy of 88.6\% at 50\% data size.

In this work, we investigate the development of a 2 step text augmentation method in order to be improve
classification models for Setswana and Sepedi. To do this we had to first identify a suitable data source. Collect
the data, and then annotate the datasets with news categories. After the data is collected and annotated, we then worked
to create classification models from the data as is and then use a word embedding and document embedding augmentation approach.

\section{Developing news classification models for Setswana and Sepedi}\label{sec:approach}

Here we discuss how data was collected as well as the approach we use to build classification models.
\subsection{Data Collection, Cleaning and Annotation}

Before we can train classification models, we first have to collect data for 2 distinct processes. First, we present our collected news dataset as well as its annotation. We then discuss how we collected larger datasets for better vectorization.

\subsubsection{News data collection and annotation}
The news data we collected is from the SABC\footnote{http://www.sabc.co.za/} Facebook pages. The SABC is the public broadcaster for South Africa. Specifically, data was collected from the Thobela FM (An SABC Sepedi radio station)\footnote{https://www.facebook.com/thobelafmyaka/} and Motsweding FM (An SABC Setswana radio station)\footnote{https://www.facebook.com/MotswedingFM/}. We scraped the news headlines that are published as posts on both Facebook pages. We claim no copyright for the content but used the data for research purposes. We summarize the datasets in Table \ref{tab:datasets}. We visualize the token distributions in Sepedi and Setswana in Figures 1 and 2 respectively.

\begin{table}[ht]
\centering
\caption{News Data Sets}
\label{tab:datasets}
\begin{tabular}{|l|l|l|} 
\hline
 & Setswana  & Sepedi \\ 
\hline
Corpus Size (headlines) & 219 & 491 \\ 
\hline
Number of Tokens (words) & 1561 & 3018 \\ 
\hline
\end{tabular}
\end{table}

\begin{figure}[ht]
\begin{center}
\includegraphics[width=\columnwidth]{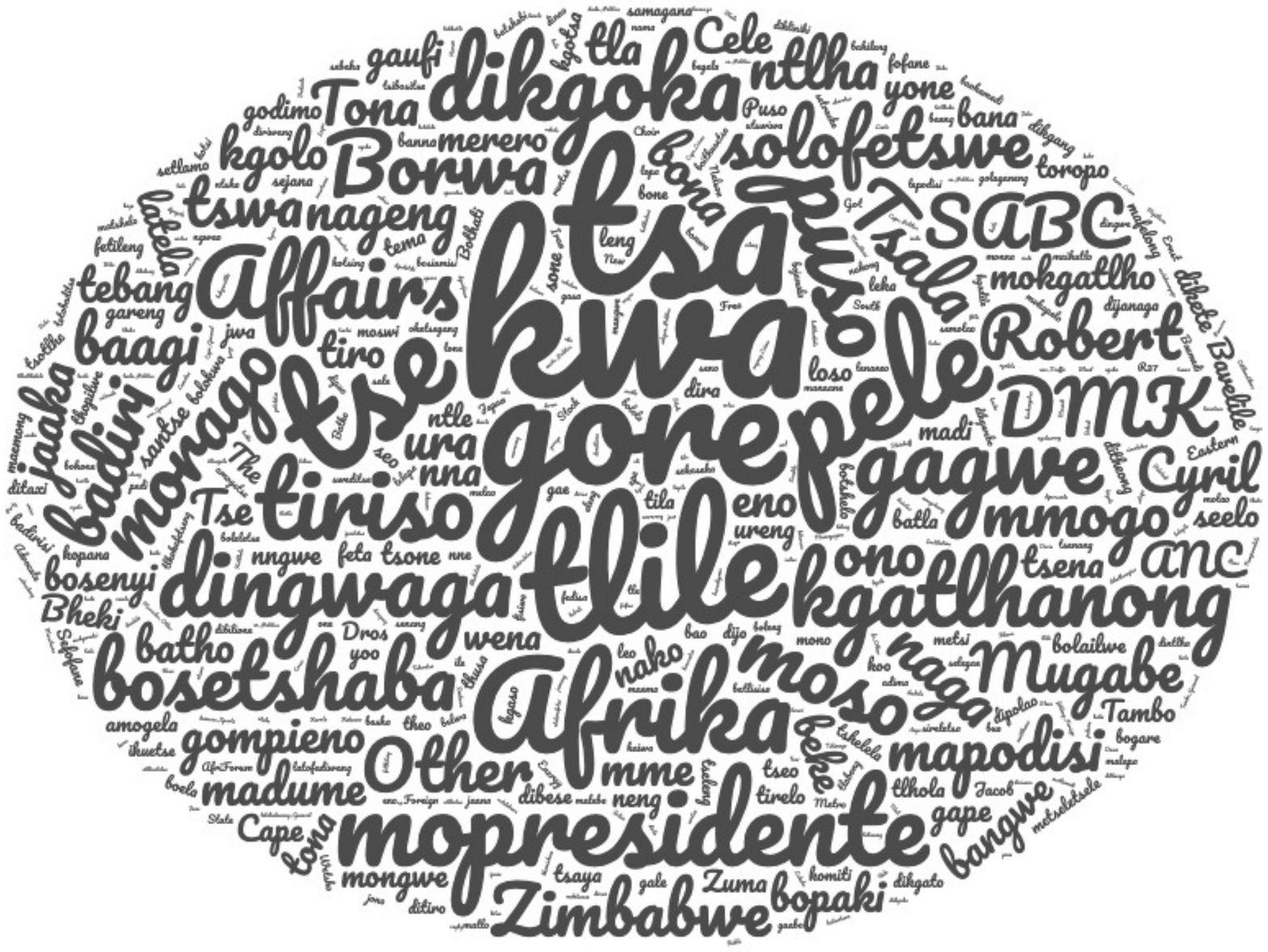} 
\label{fig.1}
\caption{Setswana Wordcloud}
\end{center}
\end{figure}

\begin{figure}[ht]
\begin{center}
\includegraphics[width=\columnwidth]{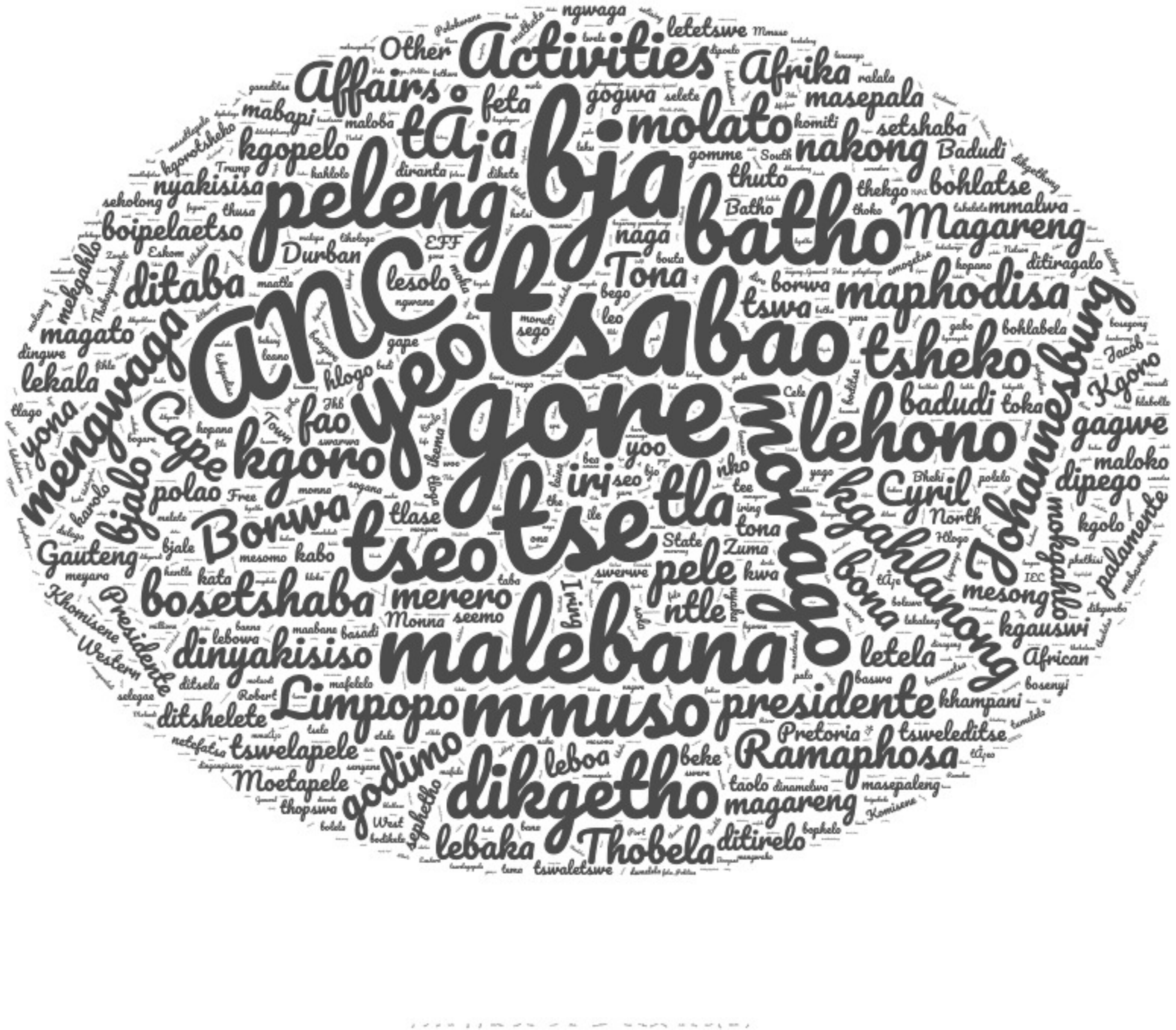} 
\label{fig:wordcloud_sepedi}
\caption{Sepedi Wordcloud}
\end{center}
\end{figure}

As can be seen, the datasets are relatively small and as such, we have to look at other ways to build vectorizers that can better generalize as the word token diversity would be very low. 

We annotated the datasets by categorizing the news headlines into: \emph{Legal}, \emph{General News},\emph{Sports}, \emph{Other}, \emph{Politics}, \emph{Traffic News}, \emph{Community Activities}, \emph{Crime}, \emph{Business} and \emph{Foreign Affairs}. Annotation was done after reading the headlines and coming up with categories that fit both datasets. We show the distribution of the labels in both the Setswana and Sepedi data sets in Figures 3 and 4 respectively. For this work, we only explore single label categorization for each article. It remains future work to look at the multi-label case. As such, there might be some noise in the labels. Examples from the Sepedi annotated news corpus are shown next:

\begin{quote}
    \emph{Tsela ya N1 ka Borwa kgauswi le Mantsole Weighbridge ka mo Limpopo ebe e tswaletswe lebakanyana ka morago ga kotsi yeo e hlagilego.} \textbf{Traffic}
    \\
    \\
    \emph{Tona ya toka Michael Masutha,ore bahlankedi ba kgoro ya ditirelo tsa tshokollo ya bagolegwa bao ba tateditswego dithieeletsong tsa khomisene ya go nyakisisa mabarebare a go gogwa ga mmuso ka nko,ba swanetse go hlalosa gore ke ka lebaka la eng ba sa swanelwa go fegwa mesomong} \textbf{Legal}
\end{quote}

\begin{figure}[ht]
\begin{center}
\includegraphics[width=\columnwidth]{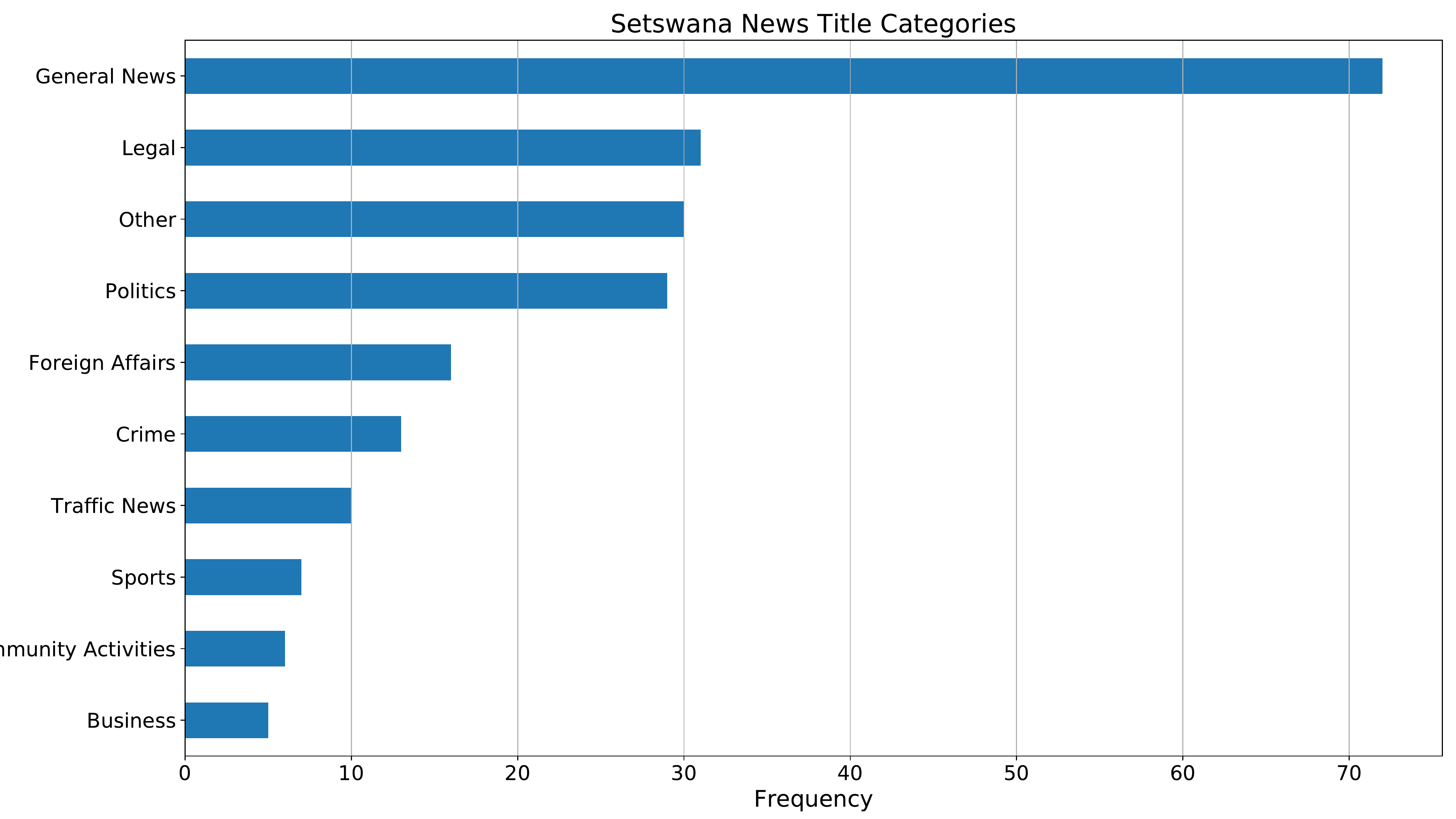} 
\label{fig:setswana_categories}
\caption{Setswana news title category distribution}
\end{center}
\end{figure}

\begin{figure}[ht]
\begin{center}
\includegraphics[width=\columnwidth]{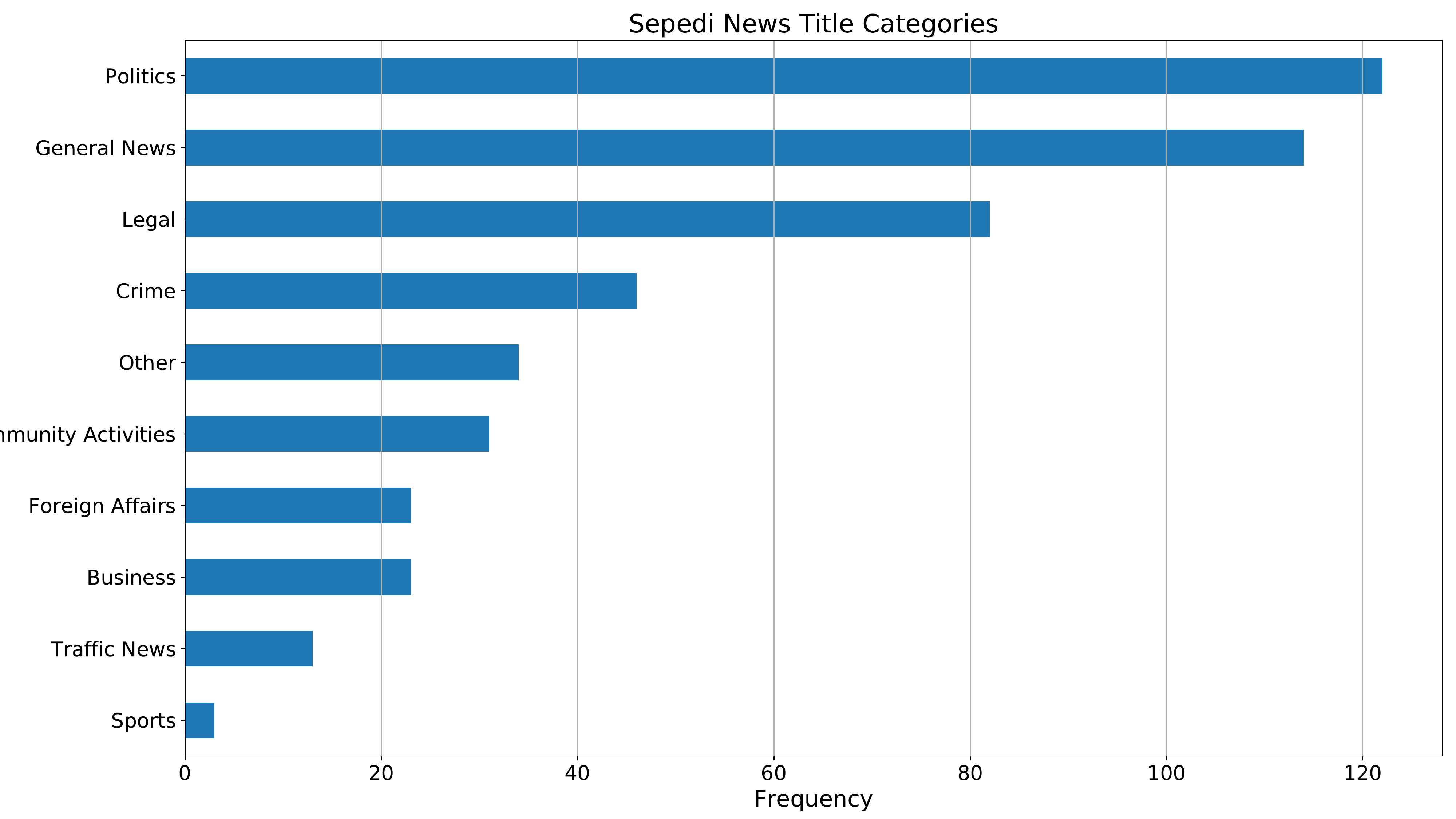} 
\label{fig:sepedi_categories}
\caption{Sepedi news title category distribution}
\end{center}
\end{figure}

The full dataset is made available online \cite{marivate_vukosi_2020_3668489} for further research use and improvements to the annotation\footnote{\url{https://zenodo.org/record/3668495}}. As previously discussed, we used larger corpora to create language vectorizers for downstream NLP tasks. We discuss this next.

\subsubsection{Vectorizers}

Before we get into the annotated dataset, we needed to create pre-trained vectorizers in order to be able to build more classifiers that generalize better later on. For this reason we collected different corpora for each language in such as way that we could create Bag of Words, TFIDF, Word2Vec \cite{mikolov2013distributed} and FastText \cite{bojanowski2017enriching} vectorizers (Table \ref{tab:vectorizer_data}). We also make these vectorizers available for other researchers to use. 

\begin{table}[ht]
\centering
\caption{Vectorizer Corpora Sizes in number of lines (number of tokens)}
\label{tab:vectorizer_data}
\begin{tabular}{|l|l|l|} 
\hline
Source & Setswana & Sepedi  \\ 
\hline
Wikipedia &  478(\emph{21924})\footnote{https://tn.wikipedia.org/} & 300(\emph{10190})\footnote{https://nso.wikipedia.org/}\\ 
\hline
JW300\footnote{http://opus.nlpl.eu/JW300.php} & 874464(\emph{70251}) & 618275(\emph{53004})\\ 
\hline
Bible & 3110(\emph{40497}) &  29723\\ 
\hline
Constitution\footnote{https://www.justice.gov.za/legislation/constitution/pdf.html} &  7077(\emph{3940}) & 6564(\emph{3819}) \\ 
\hline
SADILAR\footnote{https://www.sadilar.org/index.php/en/resources} & 33144(\emph{61766}) & 67036(\emph{87838}) \\ 
\hline
\textbf{Total }& \textbf{946264(\emph{152027})} & \textbf{721977(\emph{149355})} \\ 
\hline
\end{tabular}
\end{table}

\subsection{News Classification Models}
We explore the use of a few classification algorithms to train news classification models. Specifically we train
\begin{itemize}
    \item Logistic Regression, 
    \item Support Vector Classification,
    \item XGBoost, and
    \item MLP Neural Network.
\end{itemize}

To deal with the challenge of having a small amount of data on short text, we use data augmentation methods, specifically a word embedding based augmentation \cite{wang2015s}, approach that has been shown to work well on short text \cite{marivate2019improving}. We use this approach since we are not able to use other augmentation methods such as synonym based (requires developed Wordnet Synsets \cite{kobayashi2018contextual}), language models (larger corpora needed train) and back-translation (not readily available for South African languages). We develop and present the use of both word and document embeddings (as an augmentation quality check) inspired by a hierarchical approach to augmentation \cite{yu2019hierarchical}.

\section{Experiments and Results}\label{sec:results}
This Section presents the experiments and results. As this is still work in progress, we present some avenues explored in both training classifiers and evaluating them for the task of news headline classification for Setswana and Sepedi. 
\subsection{Experimental Setup}

For each classification problem, we perform 5 fold cross validation. For the bag-of-words and TFIDF vectorizers, we use a maximum token size of 20,000. For word embeddings and language embeddings we use size 50. All vectorizers were trained on the large corpora presented earlier. 

\subsubsection{Baseline Experiments}
\label{sec:baseline_experiments}

We run the baseline experiments with the original data using 5-fold cross validation. We show the performance (in terms of weighted F1 score) in the Figures 5 \& 6. We show the baseline results as \emph{orig}. For both the Bag-of-Words (TF) and TFIDF, the MLP performs very well comparatively to the other methods. In general the TFIDF performs better. 

\begin{figure}[ht]
\begin{center}
\includegraphics[width=\columnwidth]{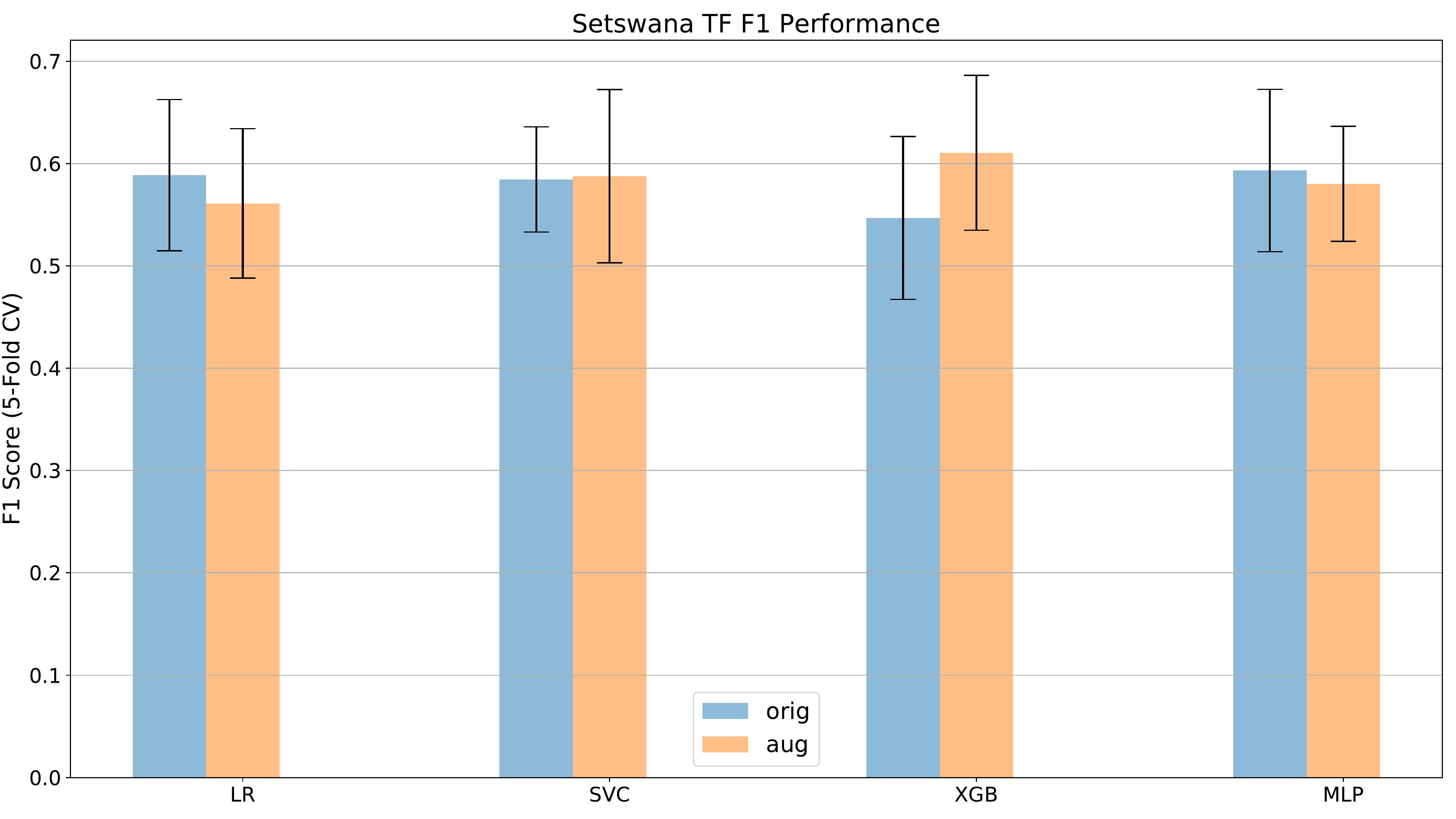} 
\includegraphics[width=\columnwidth]{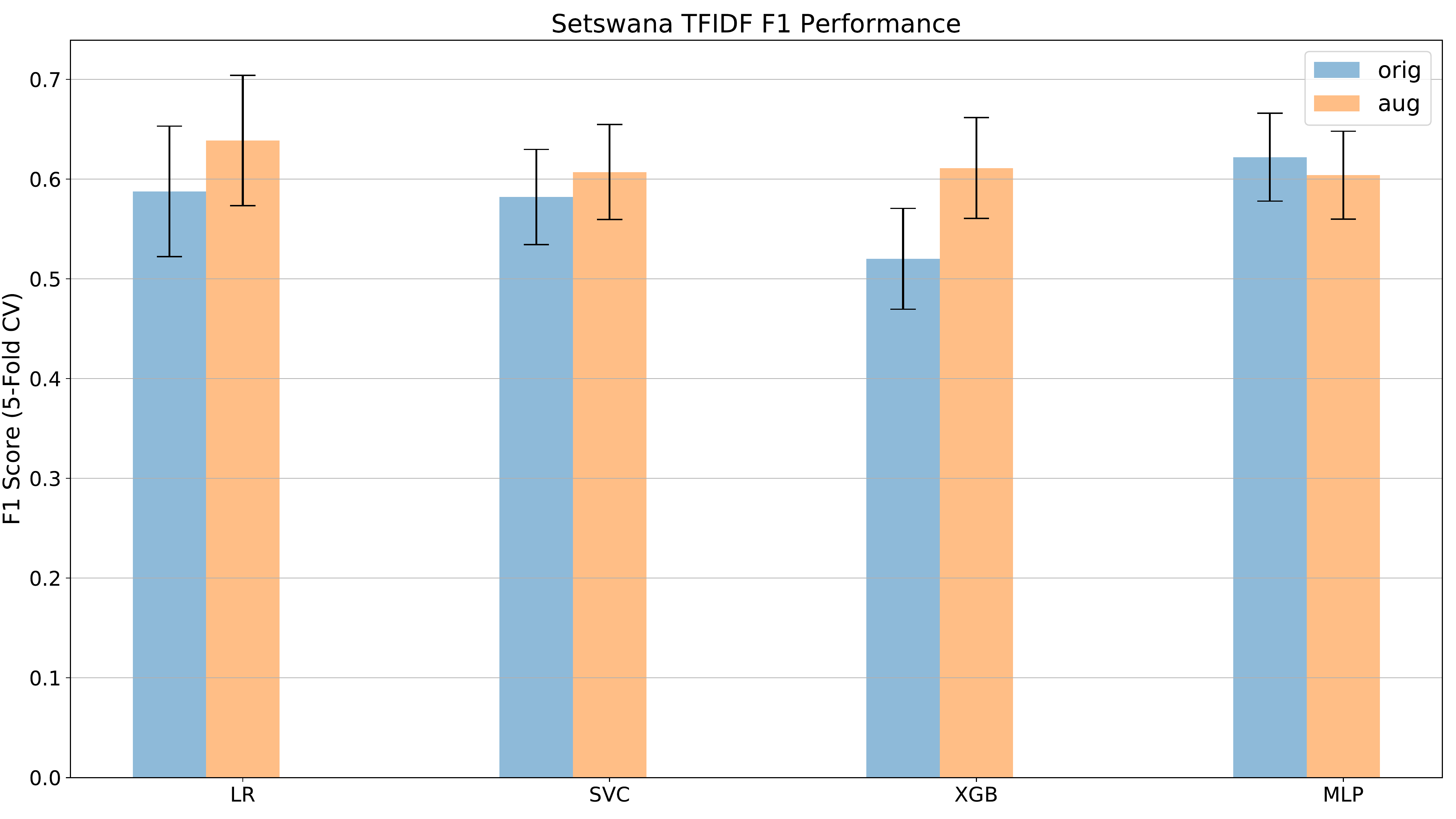} 
\label{fig:result_tf_tfidf_setswana}
\caption{Baseline classification model performance for Setswana news title categorisation}
\end{center}
\end{figure}

\begin{figure}[ht]
\begin{center}
\includegraphics[width=\columnwidth]{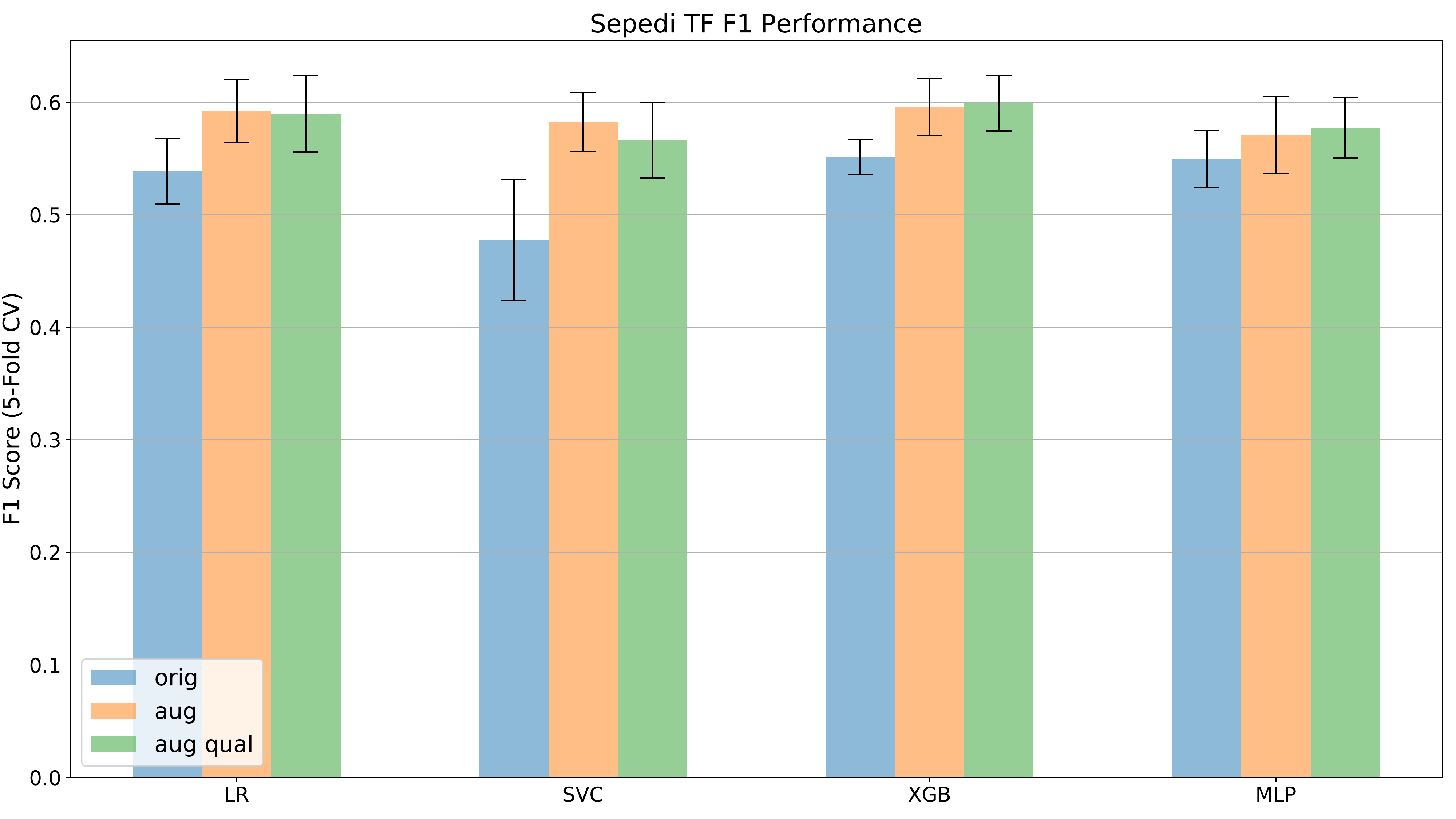} 
\includegraphics[width=\columnwidth]{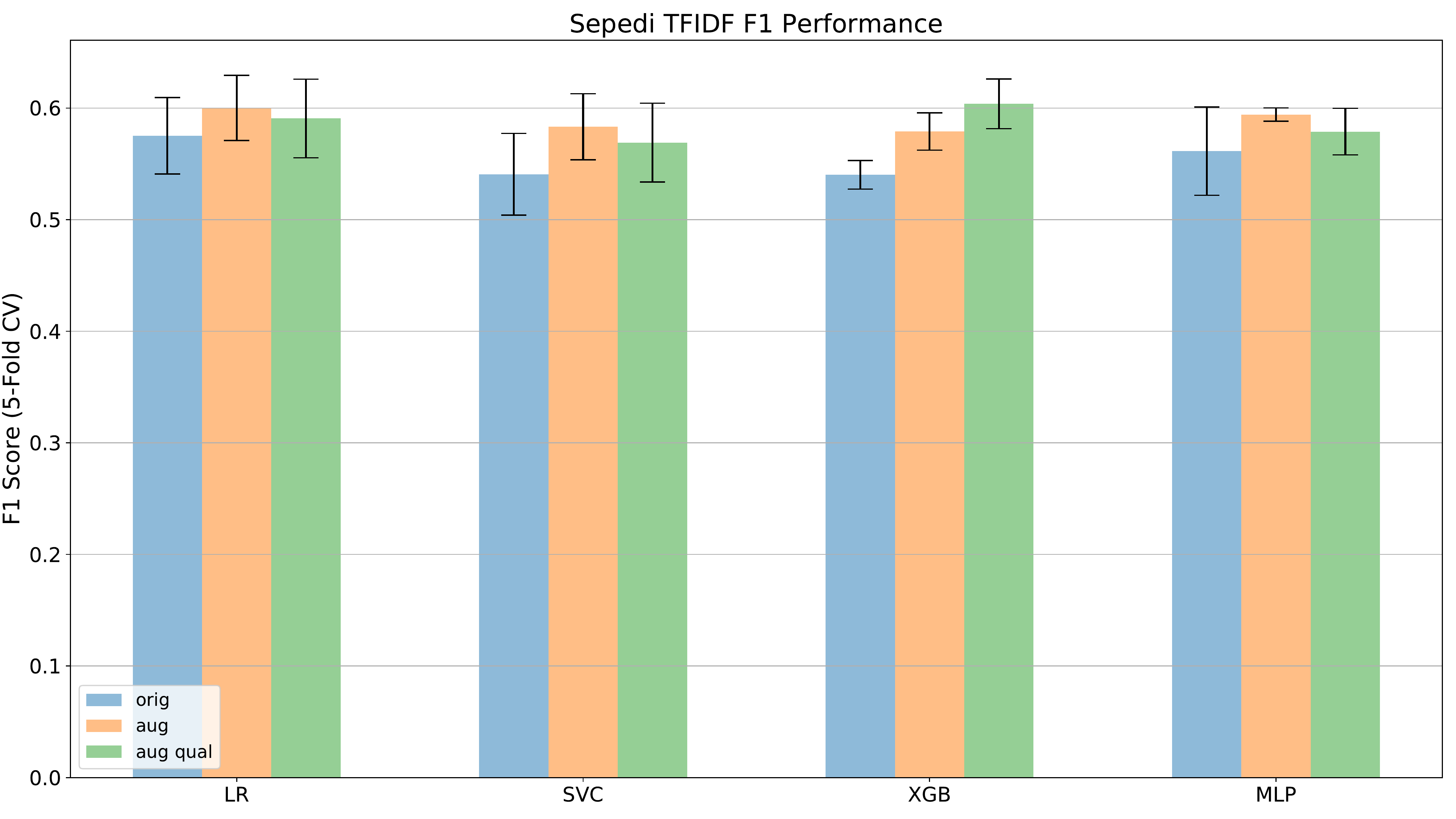} 
\label{fig:result_tf_tfidf_sepedi}
\caption{Baseline classification model performance for Sepedi news title categorisation}
\end{center}
\end{figure}
\subsubsection{Augmentation}

We applied augmentation in different ways. First for Sepedi and Setswana word embeddings (word2vec), we use word embedding-based augmentation. We augment each dataset 20 times on the training data while the validation data is left intact so as to be comparable to the earlier baselines. We show the effect of augmentation in Figure 5 and 6 (performance labeled with \emph{aug} 

The contextual, word2vec based, word augmentation improves the performance of most of the classifiers. If we now introduce a quality check using doc2vec (Algorithm 1) we also notice the impact on the performance for Sepedi (Figure 6 \emph{aug qual} ). We were not able to complete experiments with Setswana for the contextual augmentation with a quality check, but will continue working to better under stand the impact of such an algorithm in general. For example, it remains further work to investigate the effects of different similarity thresholds for the algorithm on the overall performance, how such an algorithm works on highly resourced languages vs low resourced languages, how we can make the algorithm efficient etc.

\begin{algorithm}
\SetAlgoLined
\SetKwFunction{Aug}{Augment}
\SetKwProg{Fn}{def}{\string:}{}
\SetKwFunction{Range}{range}
\SetKw{KwTo}{in}
\SetKwFor{For}{for}{\string:}{}%
\SetKwIF{If}{ElseIf}{Else}{if}{:}{elif}{else:}{}%
\SetKwFor{While}{while}{:}{fintq}%
\KwIn{$s$: a sentence, $run$: maximum number of attempts at augmentation}
\KwOut{$\hat{s}$ a sentence with words replaced}
\Fn{\Aug{s,run}}{
 Let $\vv{V}$ be a vocabulary\;
 \For{ $i$ \KwTo \Range{run} }{
    $w_i \gets$ randomly select a word from $s$\;
    $\vv{w} \gets$ find similar words of $w_i$\;
    $s_0 \gets$ randomly select a word from $\vv{w}$ given weights as distance\;
    $\hat{s} \gets $replace $w_i$ with similar word $s_0$\;
    $\vv{s} \gets Doc2vec(s)$\;
    $\vv{\hat{s}} \gets Doc2vec(\hat{s})$\;
    $similarity$ $\gets$ Cosine Similarity($\vv{s}$, $\vv{\hat{s}}$)\;
    \If {$similarity$ $>$ $threshold$} {
        return($\hat{s}$)\;
    } 
 }

}
\caption{Contextual (Word2vec-based) augmentation algorithm with a doc2vec quality check\label{word2vecqual} }
\end{algorithm}

\begin{figure}[ht]
\begin{center}
\includegraphics[width=\columnwidth]{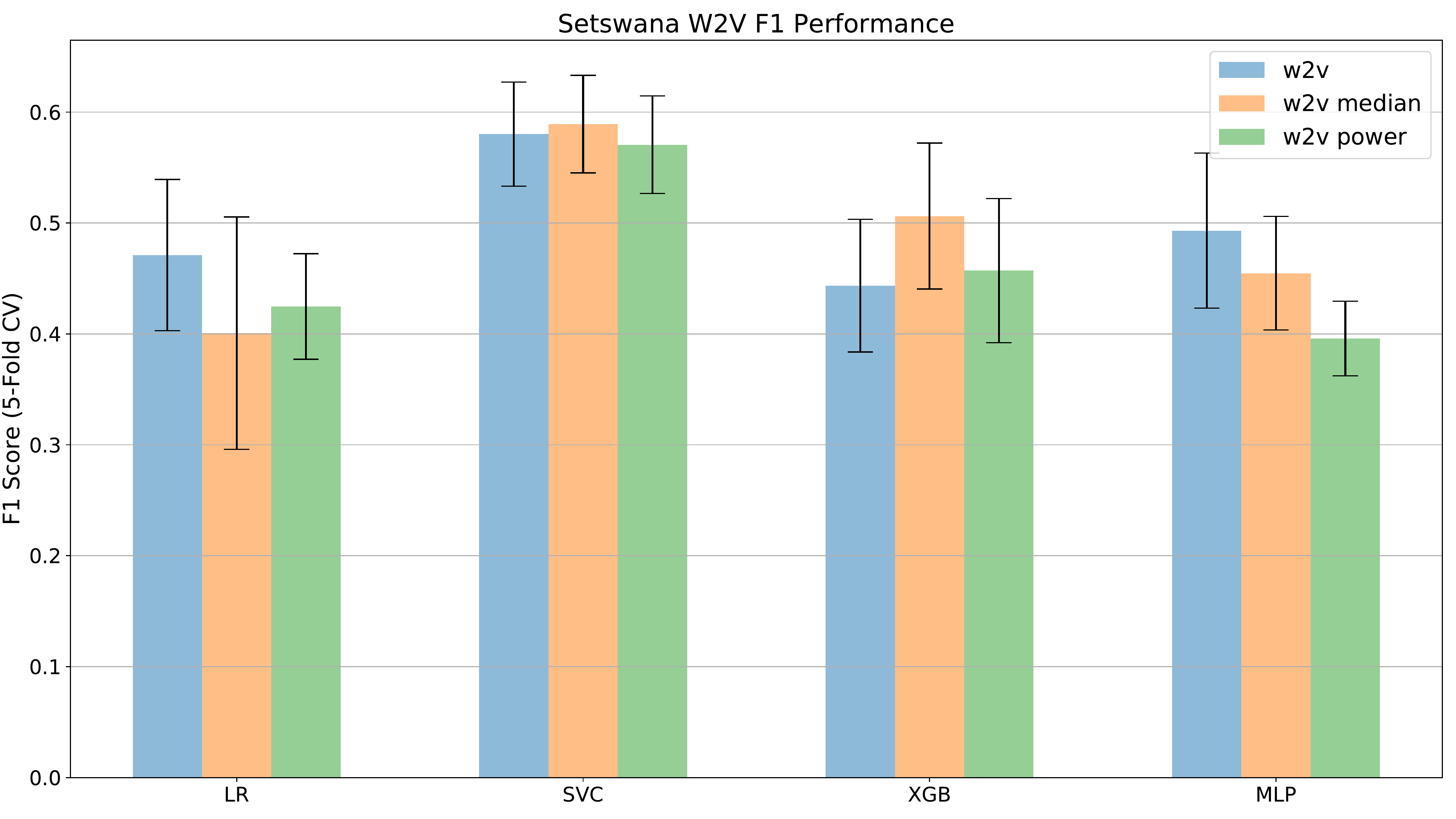} 
\includegraphics[width=\columnwidth]{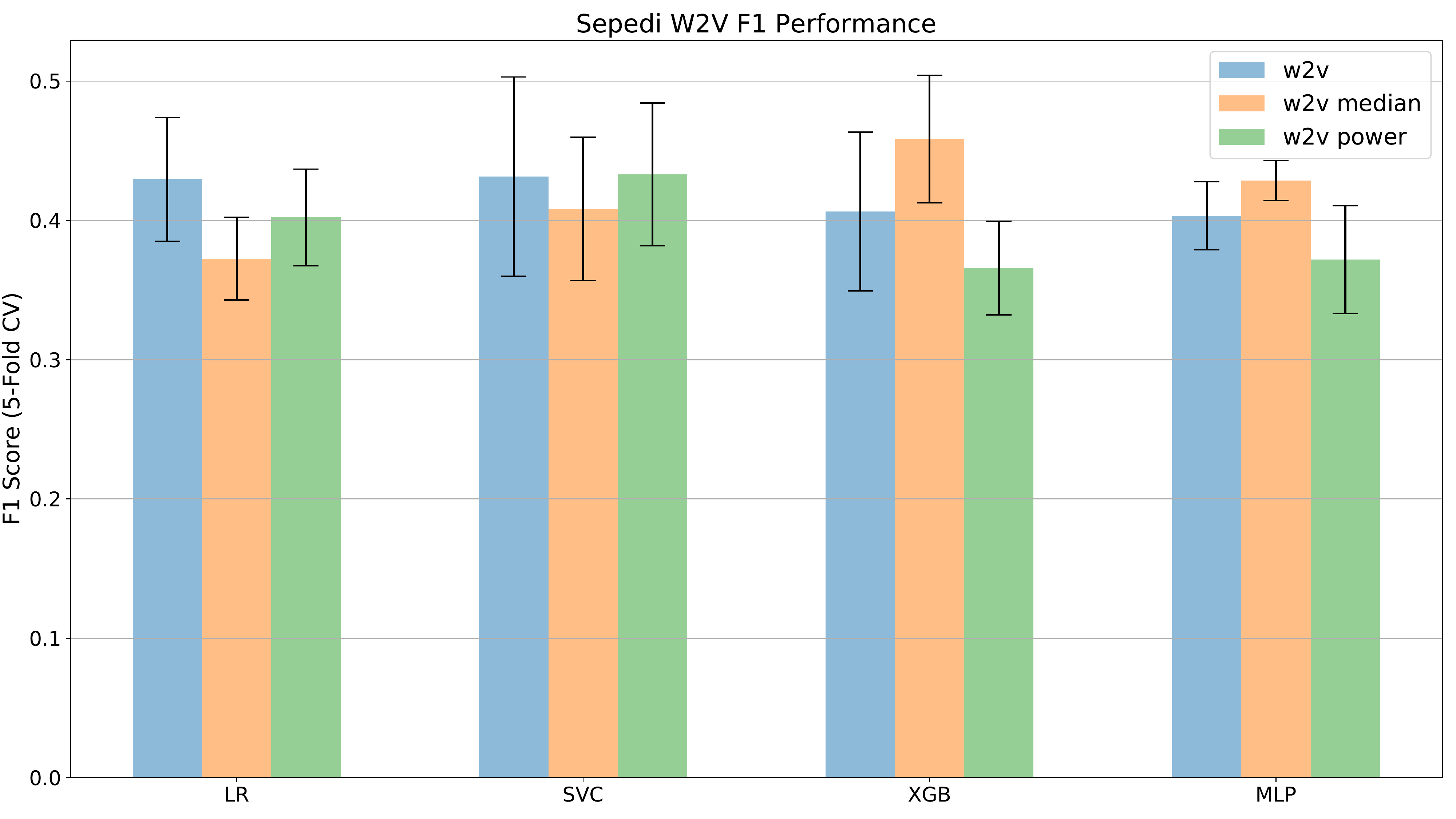}
\label{fig:result_w2v}
\caption{Word2Vec feature based performance for news headline classification}
\end{center}
\end{figure}

\begin{figure}[h]
\begin{center}
\includegraphics[width=\columnwidth]{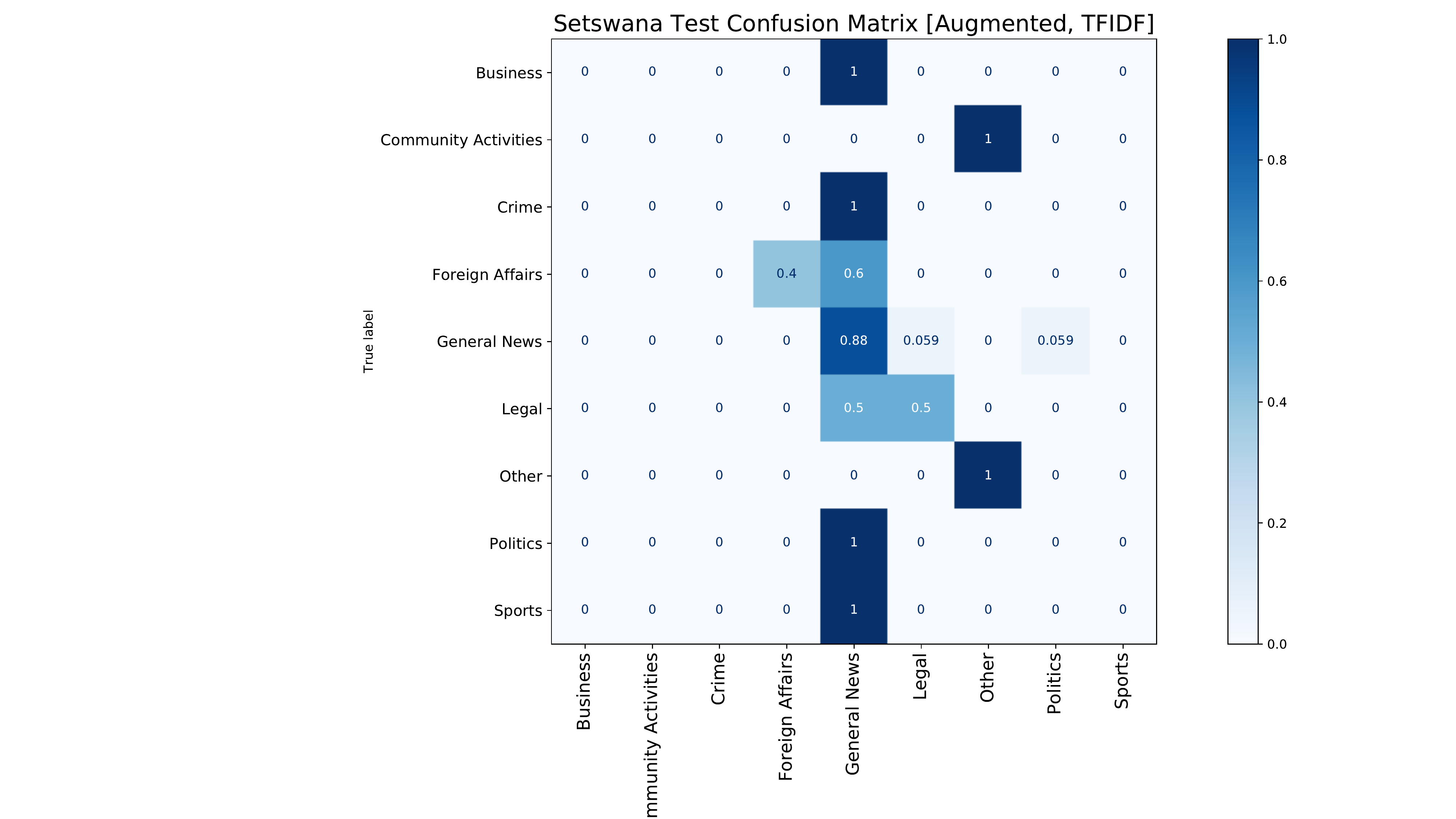} 
\includegraphics[width=\columnwidth]{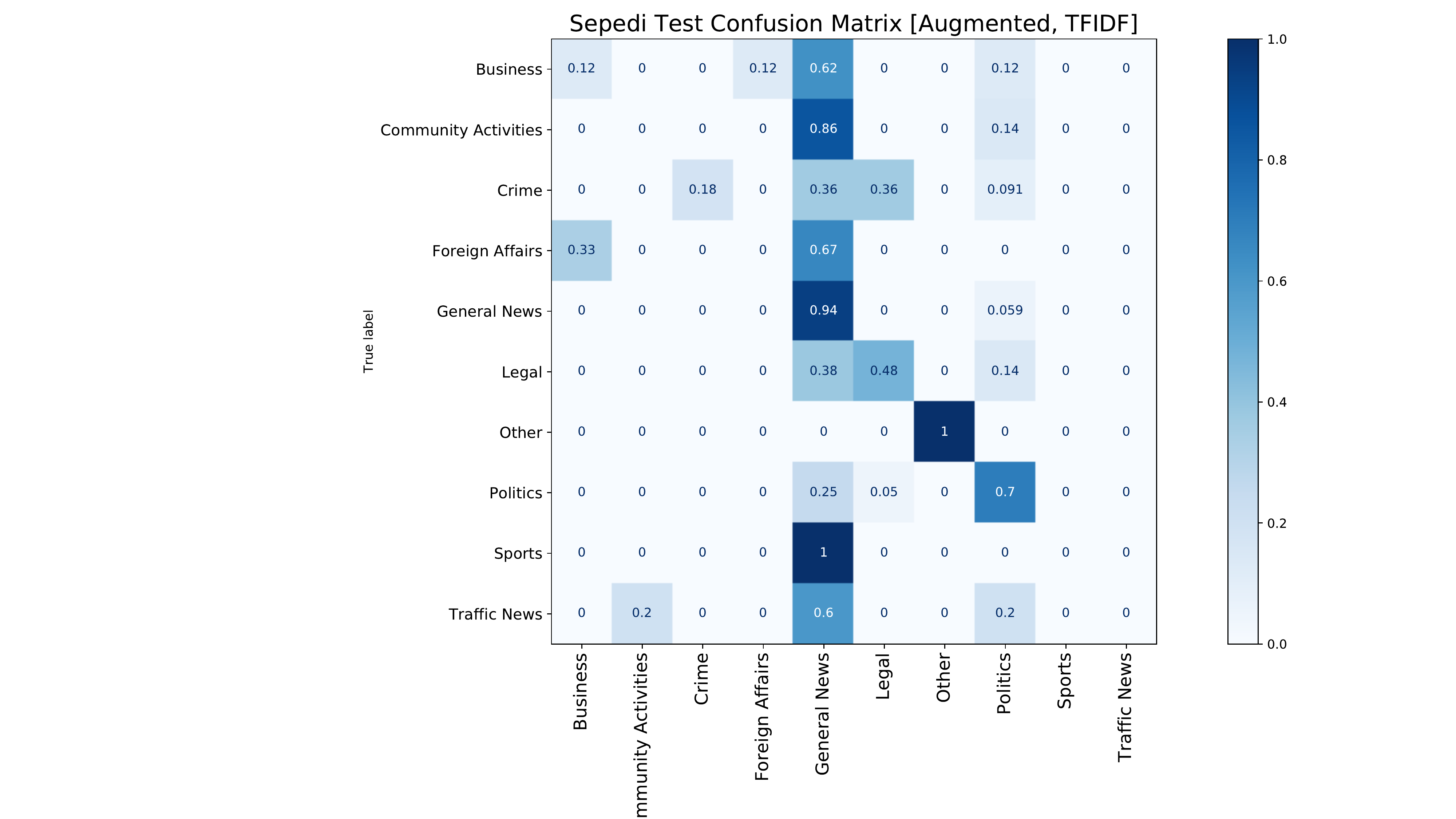} 
\label{fig:result_tf_tfidf_sepedi_aug}
\caption{Confusion Matrix of News headline classification models}
\end{center}
\end{figure}

It also interesting to look at how performance of classifiers that were only trained with word2vec features would fair. Deep neural networks are not used in this current work and as such we did not use recurrent neural networks, but we can create sentence features from - word2vec by either using: the mean of all word vectors in a sentence, the median of all word vectors in a sentence or the concatenated power means \cite{ruckle2018concatenated}. We show the performance of using this approach with the classifiers used for Bag of Words and TFIDF earlier in Figure~\ref{fig:result_tf_tfidf_sepedi_aug}. 

The performance for this approach is slightly worse with the best results for Sepedi news headline classification being with XGBoost on the augmented data. We hope to improve this performance using word2vec feature vectors using recurrent neural networks but currently are of the view that increasing the corpora sizes and the diversity of corpora for the pre-trained word embeddings may yield even better results. 

Finally, we show the confusion matrix of the best model in Sepedi on a test set in Figure 8. The classifier categorises \emph{General News}, \emph{Politics} and \emph{Legal} news headlines best. For the others there there is more error. A larger news headline dataset is required and classification performance will also need to be compared to models trained on full news data (with the article body). For the Setswana classifiers, the confusion matrix shows that the data skew results in models that mostly can categorise between categorises \emph{General News} and \emph{Other}. We need to look at re-sampling techniques to improve this performance as well as increasing the initial dataset size.

\section{Conclusion and Future Work}\label{sec:future}

This work introduced the collection and annotation of Setswana and Sepedi news headline data. It remains a challenge
that in South Africa, 9 of the 11 official languages have little data such as this that is available to researchers in order 
to build downstream models that can be used in different applications. Through this work we hope to provide an example of 
what may be possible even when we have a limited annotated dataset. We exploit the availability of other free text data in Setswana and Sepedi in order to build pre-trained vectorisers for the languages (which are released as part of this work) and then train classification models for news categories.

It remains future work to collect more local language news headlines and text to train more models. We have identified other government news sources that can be used. On training embedding models with the data we have collected, further studies are needed to look at how augmentation using the embedding models improve the quality of augmentation. 
\section{Bibliographical References}\label{reference}

\bibliographystyle{lrec}
\bibliography{references}

\begin{thebibliography}{}

\bibitem[\protect\citename{Bojanowski \bgroup et al.\egroup
  }2017]{bojanowski2017enriching}
Bojanowski, P., Grave, E., Joulin, A., and Mikolov, T.
\newblock (2017).
\newblock Enriching word vectors with subword information.
\newblock {\em Transactions of the Association for Computational Linguistics},
  5:135--146.

\bibitem[\protect\citename{{B}ureau~of {C}irculations}2019]{abc2019q1}
{B}ureau~of {C}irculations, A.
\newblock (2019).
\newblock Newspaper circulation statistics for the period january–march 2019
  (abc q1 2019).

\bibitem[\protect\citename{Cubuk \bgroup et al.\egroup
  }2019]{cubuk2019autoaugment}
Cubuk, E.~D., Zoph, B., Mane, D., Vasudevan, V., and Le, Q.~V.
\newblock (2019).
\newblock Autoaugment: Learning augmentation strategies from data.
\newblock In {\em Proceedings of the IEEE conference on computer vision and
  pattern recognition}, pages 113--123.

\bibitem[\protect\citename{Johnson \bgroup et al.\egroup
  }2017]{johnson2017google}
Johnson, M., Schuster, M., Le, Q.~V., Krikun, M., Wu, Y., Chen, Z., Thorat, N.,
  Vi{\'e}gas, F., Wattenberg, M., Corrado, G., et~al.
\newblock (2017).
\newblock Google’s multilingual neural machine translation system: Enabling
  zero-shot translation.
\newblock {\em Transactions of the Association for Computational Linguistics},
  5:339--351.

\bibitem[\protect\citename{Kobayashi}2018]{kobayashi2018contextual}
Kobayashi, S.
\newblock (2018).
\newblock Contextual augmentation: Data augmentation by words with paradigmatic
  relations.
\newblock In {\em Proceedings of the 2018 Conference of the North American
  Chapter of the Association for Computational Linguistics: Human Language
  Technologies, Volume 2 (Short Papers)}, volume~2, pages 452--457.

\bibitem[\protect\citename{Lewis}1997]{lewis1997reuters}
Lewis, D.~D.
\newblock (1997).
\newblock Reuters-21578 text categorization collection data set.

\bibitem[\protect\citename{Marivate and Sefara}2019]{marivate2019improving}
Marivate, V. and Sefara, T.
\newblock (2019).
\newblock Improving short text classification through global augmentation
  methods.
\newblock {\em arXiv preprint arXiv:1907.03752}.

\bibitem[\protect\citename{Marivate and
  Sefara}2020a]{vukosi_marivate_2020_3668481}
Marivate, V. and Sefara, T.
\newblock (2020a).
\newblock African embeddings [nlp].
\newblock \url{https://doi.org/10.5281/zenodo.3668481}, February.

\bibitem[\protect\citename{Marivate and
  Sefara}2020b]{marivate_vukosi_2020_3668489}
Marivate, V. and Sefara, T.
\newblock (2020b).
\newblock South {A}frican news data dataset.
\newblock \url{https://doi.org/10.5281/zenodo.3668489}.

\bibitem[\protect\citename{Mikolov \bgroup et al.\egroup
  }2013]{mikolov2013distributed}
Mikolov, T., Sutskever, I., Chen, K., Corrado, G.~S., and Dean, J.
\newblock (2013).
\newblock Distributed representations of words and phrases and their
  compositionality.
\newblock In {\em Advances in neural information processing systems}, pages
  3111--3119.

\bibitem[\protect\citename{Nettle}1998]{nettle1998explaining}
Nettle, D.
\newblock (1998).
\newblock Explaining global patterns of language diversity.
\newblock {\em Journal of anthropological archaeology}, 17(4):354--374.

\bibitem[\protect\citename{R{\"u}ckl{\'e} \bgroup et al.\egroup
  }2018]{ruckle2018concatenated}
R{\"u}ckl{\'e}, A., Eger, S., Peyrard, M., and Gurevych, I.
\newblock (2018).
\newblock Concatenated power mean word embeddings as universal cross-lingual
  sentence representations.
\newblock {\em arXiv preprint arXiv:1803.01400}.

\bibitem[\protect\citename{Sandhaus}2008]{sandhaus2008new}
Sandhaus, E.
\newblock (2008).
\newblock The new york times annotated corpus.
\newblock {\em Linguistic Data Consortium, Philadelphia}, 6(12):e26752.

\bibitem[\protect\citename{Silfverberg \bgroup et al.\egroup
  }2017]{silfverberg2017data}
Silfverberg, M., Wiemerslage, A., Liu, L., and Mao, L.~J.
\newblock (2017).
\newblock Data augmentation for morphological reinflection.
\newblock In {\em Proceedings of the CoNLL SIGMORPHON 2017 Shared Task:
  Universal Morphological Reinflection}, pages 90--99.

\bibitem[\protect\citename{Strassel and Tracey}2016]{strassel2016lorelei}
Strassel, S. and Tracey, J.
\newblock (2016).
\newblock Lorelei language packs: Data, tools, and resources for technology
  development in low resource languages.
\newblock In {\em Proceedings of the Tenth International Conference on Language
  Resources and Evaluation (LREC'16)}, pages 3273--3280.

\bibitem[\protect\citename{Wang and Yang}2015]{wang2015s}
Wang, W.~Y. and Yang, D.
\newblock (2015).
\newblock That's so annoying!!!: A lexical and frame-semantic embedding based
  data augmentation approach to automatic categorization of annoying behaviors
  using\# petpeeve tweets.
\newblock In {\em Proceedings of the 2015 Conference on Empirical Methods in
  Natural Language Processing}, pages 2557--2563.

\bibitem[\protect\citename{Wei and Zou}2019]{wei2019eda}
Wei, J. and Zou, K.
\newblock (2019).
\newblock Eda: Easy data augmentation techniques for boosting performance on
  text classification tasks.
\newblock In {\em Proceedings of the 2019 Conference on Empirical Methods in
  Natural Language Processing and the 9th International Joint Conference on
  Natural Language Processing (EMNLP-IJCNLP)}, pages 6383--6389.

\bibitem[\protect\citename{Yu \bgroup et al.\egroup }2019]{yu2019hierarchical}
Yu, S., Yang, J., Liu, D., Li, R., Zhang, Y., and Zhao, S.
\newblock (2019).
\newblock Hierarchical data augmentation and the application in text
  classification.
\newblock {\em IEEE Access}, 7:185476--185485.

\bibitem[\protect\citename{Zhang \bgroup et al.\egroup
  }2015]{zhang2015character}
Zhang, X., Zhao, J., and LeCun, Y.
\newblock (2015).
\newblock Character-level convolutional networks for text classification.
\newblock In {\em Advances in neural information processing systems}, pages
  649--657.

\bibitem[\protect\citename{Zoph \bgroup et al.\egroup }2016]{zoph2016transfer}
Zoph, B., Yuret, D., May, J., and Knight, K.
\newblock (2016).
\newblock Transfer learning for low-resource neural machine translation.
\newblock {\em arXiv preprint arXiv:1604.02201}.

\end{thebibliography}


\end{document}